# OPTİMİZE EDİLMİŞ ÇKA İLE COVID-19 SINIFLANDIRMASI İÇİN KAYNAŞTIRILMIŞ DERİN ÖZELLİKLERE DAYALI SINIFLANDIRMA ÇERÇEVESİ

[1]Şaban ÖZTÜRK, [2]Enes YİĞİT, [3]Umut ÖZKAYA

[1]*Amasya Üniversitesi, Teknoloji Fakültesi, Elektrik-Elektronik Mühendisliği Bölümü, AMASYA*
[2]*Karamanoğlu Mehmetbey Üniversitesi, Mühendislik ve Mimarlık Fakültesi, Elektrik-Elektronik Mühendisliği Bölümü, KARAMAN*
[3] *Konya Teknik Üniversitesi, Mühendislik ve Doğa Bilimleri Fakültesi, Elektrik-Elektronik Mühendisliği Bölümü, KONYA*
[1] saban.ozturk@amasya.edu.tr , [2] enesyigit@kmu.edu.tr , [3] uozkaya@ktun.edu.tr

**ÖZ:** COVID-19 adı verilen yeni tip Koronavirüs hastalığı oldukça hızlı yayılmaya devam etmektedir. Bazı spesifik semptomlar gösterse de hemen her bireyde farklı semptomlar gösterebilen bu hastalık yüzbinlerce hastanın hayatını kaybetmesine neden olmuştur. Sağlık uzmanları, daha fazla yaşam kaybını önlemek için çok çalışsalar da, hastalık yayılma oranı çok yüksektir. Bu nedenle Bilgisayar Destekli Teşhis (BDT) ve Yapay Zeka (YZ) algoritmalarının desteği hayati önem taşımaktadır. Bu çalışmada, belirtilen COVID-19 algılama ihtiyaçlarını karşılamak için günümüzün en etkili görüntü analiz yöntemi olan Evrişimli Sinir Ağı (ESA) mimarisinin optimizasyonuna dayalı bir yöntem önerilmiştir. İlk olarak, COVID-19 görüntüleri ResNet-50 ve VGG-16 mimarileri kullanılarak eğitilir. Ardından, bu iki mimarinin son katmanındaki özellikler füzyon işlemi uygulanmıştır.  Füzyon işlemi ile elde edilen bu yeni görüntü özellikleri matrisleri, COVID-19 tespiti için sınıflandırılır. Sınıflandırma işlemi için Balina Optimizasyon Algoritması (BOA) ile optimize edilmiş Çok Katmanlı Bir Algılayıcı (ÇKA) yapısı kullanılır. Elde edilen sonuçlar, önerilen çerçevenin performansının VGG-16 performansından neredeyse % 4,5 ve ResNet-50 performansından neredeyse % 3,5 daha yüksek olduğunu göstermektedir.

***Anahtar Kelimeler:*** *COVID-19, Koronavirus, Sınıflama, ÇKA, Özellik Füzyonu.*

**Fused Deep Features Based Classification Framework for COVID-19 Classification with Optimized MLP**

**ABSTRACT:** The new type of Coronavirus disease called COVID-19 continues to spread quite rapidly. Although it shows some specific symptoms, this disease, which can show different symptoms in almost every individual, has caused hundreds of thousands of patients to die. Although healthcare professionals work hard to prevent further loss of life, the rate of disease spread is very high. For this reason, the help of computer aided diagnosis (CAD) and artificial intelligence (AI) algorithms is vital. In this study, a method based on optimization of convolutional neural network (CNN) architecture, which is the most effective image analysis method of today, is proposed to fulfill the mentioned COVID-19 detection needs. First, COVID-19 images are trained using ResNet-50 and VGG-16 architectures. Then, features in the last layer of these two architectures are combined with feature fusion. These new image features matrices obtained with feature fusion are classified for COVID detection. A multi-layer perceptron (MLP) structure optimized by the whale optimization algorithm is used for the classification process. The obtained results show that the performance of the proposed framework is almost 4.5% higher than VGG-16 performance and almost 3.5% higher than ResNet-50 performance.

1. INTRODUCTION

The new type of coronavirus, called COVID-19 by the world health organization (WHO), spread rapidly among people and turned into a serious epidemic. First seen in Wuhan, China, COVID-19 spread all over the world in a short time. This virus, which seriously threatens human health, has caused the death of many people (Jaiswal *et al.*, 2020). Among the most common symptoms of this disease are fever, cough, and breathing problems. However, these symptoms and their severity differ from person to person (Öztürk *et al.*, 2020). An effective and approved vaccine for COVID-19, which can spread quite quickly through airborne droplets, has yet to be found. In addition, there is still no consensus on a definitive treatment method. As a result of all these facts, governments are trying to mitigate the epidemic by introducing serious measures and various rules. Although these measures are different from the usual social order, they begin to be accepted as new normal. In order to return to the old social order and end the COVID-19 pandemic, researchers are doing everything. Especially, researchers in the medical field carry out very devoted studies in terms of vaccines, medicines, and medical applications.

Considering the workload on medical professionals, it is clear that it is necessary to leverage technological developments to find a solution to COVID-19. Today, developments such as the widespread use of technological devices and the integration of artificial intelligence algorithms in almost every field are very promising (Vaishya *et al.*, 2020). To help combat COVID-19 by shifting these advances in technology to the medical domain, many AI researchers focus their full concentration on this area. For this purpose, researchers are working on automatic analysis of chest X-ray and CT images. Because Real Time Reverse Transcription Polymerase Chain Reaction (RT-PCR) method is time-consuming and it has error-prone results (Zu *et al.*, 2019).

CNN, the most powerful image processing method of today, is frequently used for processing X-ray and CT images related to COVID-19. In some studies for COVID-19 detection, hand-crafted features or simpler techniques are used. These methods are often preferred when there is not enough dataset to train a CNN architecture. When the COVID-19 epidemic started, some studies used hand-crafted methods because there were not enough COVID-19 datasets containing enough samples in the literature. Data sets containing insufficient samples cause overfitting problems in a deep CNN network. The most striking works performed without using any deep learning architecture are briefly summarized below. Barstugan *et al.* (2020) used Grey Level Co-occurrence Matrix (GLCM), Local Directional Pattern (LDP), Grey Level Run Length Matrix (GLRLM), Grey-Level Size Zone Matrix (GLSZM), and Discrete Wavelet Transform (DWT) algorithms to extract image features. Randhawa *et al.* (2020) proposed supervised machine learning with digital signal processing (MLDSP) for genome analyses of COVID-19. Öztürk *et al.* (2020) proposed a hybrid method that includes image augmentation and data oversampling with hand-crafted features. Elaziz *et al.* (2020) presented a machine learning method for the chest x-ray images using new Fractional Multichannel Exponent Moments (FrMEMs). Shi *et al.* (2020) proposed an infection Size Aware Random Forest method (iSARF) in order to automatically categorize images into groups. Sun *et al.* (2020) proposed an Adaptive Feature Selection guided Deep Forest (AFS-DF) for COVID-19 classification based on chest CT images. The results produced by hand-crafted methods are inspiring. However, as the number and variety of images related to COVID-19 increased, the performance of many of these methods did not meet the expectation. With the emergence of datasets containing sufficient number and variety of samples to train CNN architectures, studies containing CNN have started to emerge rapidly.

It is almost impossible to examine all of the CNN-guided COVID-19 detection studies available in the literature. Currently, the number of COVID-19 classification, segmentation, detection, etc studies is more than 50,000. For this reason, some of the most interesting and groundbreaking studies are summarized below. Hemdan *et al.* (2020) proposed a deep learning framework namely COVIDX-net. It consists seven different CNN architectures such as VGG19, MobileNet, and etc. Ozturk *et al.* (2020) proposed a linear CNN model both binary and multi-class COVID-19 image classification. Afshar *et al.* (2020) presented capsule network approach, referred to as the COVID-CAPS, being capable of handling small datasets. Their methods are based on the capsule network approach. Sahlol *et al.* (2020) combined CNN and a swarm-based feature selection algorithm to classify COVID-19 X-ray images. They facilitated the feature extraction strength of CNN and Marine Predators Algorithm to select the most relevant features. Nour *et al.* (2020) used CNN architecture to extract robust features from COVID-19 images. They feed machine learning algorithms using these deep features (k-nearest neighbor, support vector machine, and decision tree). Singh *et al.* (2020) presented a deep CNN method to classify chest X-ray-based COVID-19 images. Also, they tuned parameters of CNN using Multi-objective Adaptive Differential Evolution (MADE). Ucar and Korkmaz (2020) proposed a type of SqueezeNet architecture for classification of COVID-19 related chest X-ray images (named as COVIDiagnosis-Net). Their architecture is tuned for the COVID-19 diagnosis with Bayesian optimization additive. Fan *et al.* (2020) proposed COVID-19 Lung Infection Segmentation Deep Network (Inf-Net) for segmentation of CT slices. Wang et al. (2020) proposed a method based on a weakly-supervised to classify and localize COVID-19 lesion From CT images. Pham (2020) implemented a comprehensive study that includes pre-trained CNN models to classify COVID-19. Albahri et al. (2020) applied taxonomy analysis on binary and multi-class COVID-19 classification problems. Pereira et al (2020) proposed a CNN method with early and late fusion to analyze COVID-19 infected X-ray images.

In this study, a highly efficient AI method that takes advantage of the feature representation power of different CNN architectures is presented. VGG-16 and RecNet-50 methods, which are the two most powerful CNN methods used as backbones in the literature, are used to extract features from images. There are studies that classify the feature vectors obtained by these two architectures separately. But the thought that combining the feature representation ability of the two architectures will improve performance is quite exciting. For this purpose, feature vectors in the previous layer of the softmax layer are taken in both architectures. These feature vectors need to be combined into a single feature vector. An MLP structure is used to handle this step in an end-to-end manner. Two feature vectors are applied to the MLP input. MLP parameters are updated with the whale optimization algorithm. The most important contributions of the proposed method can be summarized as follows:

- An approach is proposed that combines the power of two different CNN architectures with powerful feature extraction capabilities.
- Robust combination of feature vectors in an end-to-end fashion using the whale optimization algorithm
- The proposed method, which is very easy to apply, significantly reduces the rate of misdiagnosis.

The rest of this study is organized as follows: Section 2 includes dataset details, methodological background, and details of the proposed method. Experimental settings and performance indicators of the proposed method is presented in Section 3. Conclusion is given in Section 4.

## 2. MATERIAL AND METHODS

SARS-CoV-2 CT scan dataset is used to test the performance of the proposed method (Soares *et al.*, 2020). It is also possible to access this data on the kaggle (https://www.kaggle.com/plameneduardo/sarscov2-ctscan-dataset). The dataset consists of 2482 CT images in total. 1252 of these images belong to patients infected by SARS-CoV-2. In 1230 CT images, it contains CT images of patients not infected with SARS-CoV-2. However, these 1230 images show CT images of other pulmonary diseases patients. For this reason, dataset is relatively challenging. We did not apply any pre-processing or augmentation to the original images in the data. This is because the performance of the proposed method can be fairly compared with the performance of other state-of-the-art methods. Some of the sample images of dataset are shown in Figure 1. Figure 1 (a) shows images of patients infected by SARS-CoV-2, Figure 1 (b) shows other images.

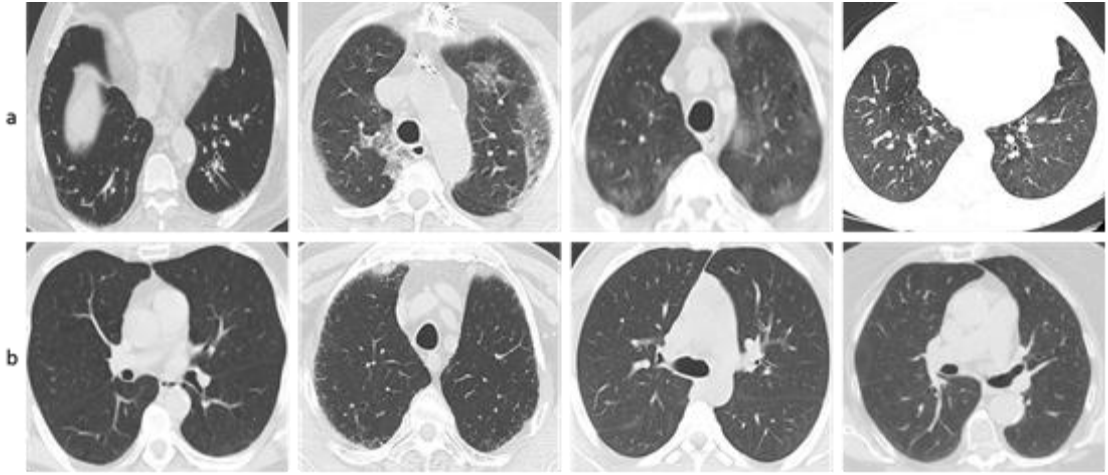

*Figure 1. Sample images from SARS-CoV-2 CT scan dataset, a) infected by SARS-CoV-2 CT scan dataset, b) non-infected by SARS-CoV-2 CT scan dataset*

CNN architectures have achieved significant success since the day it was first proposed. By solving many image processing problems, it has become a milestone in image analysis. Of course, It did not realize all these with just one architecture. New CNN architectures have been suggested frequently since its inception. For this reason, it is possible to find many different CNN architectures in the literature. Some of these architectures are very popular, while others are almost never used. In this study, VGG-16 and ResNet-50 architectures, which are accepted by artificial intelligence researchers and used as backbones in many studies, are used. Before going into the details of these architectures, a brief information about the main CNN layers will be useful. CNN architectures generally consist of specific layers and various connection types. Convolution layer is the layers where learned properties are stored. It can be said that this layer, which contains a single kernel consisting of two-dimensional matrices, is the most important layer of a CNN architecture. The kernels in this layer are scrolled on the image. This enables weight sharing and spatial information features. The other basic layer is the pooling layer. The pooling layer reduces the image dimensions while preserving the important features in the image. Thus, the number of parameters that need to be trained in the CNN architecture is greatly reduced. It has types such as max-pooling, average-pooling, sum-pooling. ReLU is the activation function. It is a very fast and simple function used to break the linearity in the network. Fully connected layer (FCL) is actually a kind of ANN structure. In this section, matrices are transformed into vectors and processed with the help of neurons. Softmax layer is preferred for classifying the vectors at the output according to their ratios. Finally, if a residual architecture is to be used (eg ResNet), the concatenate layer is required. This layer combines different layer outputs (Öztürk and Özkaya, 2020). After this quick and basic introduction to CNN layers, it will be useful to calculate a CNN output consisting of a convolution, pooling and ReLU layer. Equation 1 is used for this operation.

$$f(l_{next}) = pool_{nxn}\left(\sigma\left(w \otimes [D_{in}] + b\right)\right) \tag{1}$$

in which, $f$ represents CNN, $l_{next}$ is the input of the next layer of the output of the current layer, *pool* represents pooling layer (max-pooling, average-pooling, or sum-pooling), *nxn* represents pooling window, $\sigma$ is ReLU function, $w$ represents convolution layer, $D_{in}$ is the input of the current layer, $b$ is the bias value.

VGG-16 architecture (Simonyan and Zisserman, 2014). includes 16 weight layers. So it contains 13 convolution layers and 3 FCL layers. Parameters in these layers are updateable. All convolution layers are 3x3 pixels with stride of 1. All pooling layers consist of 2x2 windows with stride of 2. Two FCLs consist of 4,096 nodes while the last FCL consists of 1000 nodes. It includes nearly 138 million trainable parameters.

ResNet-50 architecture (He *et al.*, 2016) is a very striking structure consisting of residual modules. It includes 50 weight layers. Although it is deeper than the VGG-16 model, it takes up less space. It consists of five stages each with a convolution and identity block. Each convolution block has three convolution layers and each identity block also has 3 convolution layers. It has nearly 23 million trainable parameters.

An MLP structure is a type of the feedforward ANN architecture. A basic MLP structure consists of at least three layers: an input layer, a hidden layer, and an output layer. It is possible to increase the depth of MLP architecture. For this, the number of hidden layers in the MLP structure is increased. Neurons in MLP structure are generally followed by a nonlinear activation function. It is possible to train MLP architecture with backpropagation methods. Although it is a type of ANN, in some cases the name MLP is used in structures consisting of multiple layers of perceptrons. The selection of activation functions is one of the most important steps in solving a problem. In addition, it is very important in algorithms used for updating trainable parameters. Stochastic gradient descent (SGD) is widely used for the optimization of parameters (Wu *et al.*, 2020). In the optimization process performed with SGD, it may take time to reach global minima, and problems with vanishing of gradients or explosion of gradients may be encountered. In some cases, it remains stuck in the local minima. Many optimization techniques have been proposed to overcome these problems. In this study, MLP parameters are updated with the whale optimization algorithm (WOA) (Mirjalili and Lewis, 2016). WOA is a meta-heuristic optimization algorithm. It is based on the hunting strategy of humpback whales. For a brief information about the WOA algorithm, the encircling prey, spiral bubble-net feeding maneuver, and search for prey stages are mathematically examined. In the encircling prey stage, humpback whales determine the location of the prey and surround the prey. The WOA algorithm tries to determine the optimum position for the hunt. After the best search agent is determined, other search agents update their positions. The purpose of this update process is to move towards the best location. Equations 2 and 3 define this process.

$$D = |C.X^*(t) - X(t)| \tag{2}$$

$$X(t+1) = X^*(t) - A.D \tag{3}$$

where *t* represents the current iteration, *C* and *A* indicate coefficient vectors, $X^*$ represents the best solution position vector, *X* indicates the current position vector, and '.' represents element-by-element multiplication. Also, other terms are calculated; $A=2a.r-a$ and $C=2.r$.

Two different approaches named as 'shrinking encircling mechanism' and 'spiral updating position' are used to better define the bubble-net behavior. In the shrinking encircling mechanism, the area is narrowed by decreasing the value of the *a* parameter. In this case the *A* is a random value in the interval [-*a*, *a*]. In the spiral updating position stage, the distance between the whale's location and the prey location is calculated. A spiral equation is defined between these two locations. With this equality, the whale updates its position with a helix-shaped movement. In Equation 4 this movement is defined.

$$X(t+1) = |X^*(t) - X(t)|.e^{bl}.\cos(2\pi l) + X^*(t) \tag{4}$$

in which, *b* defines shape of the logarithmic spiral, l represents random number in the interval [-1, 1]. In search for prey (exploration) phase, humpback whales take positions at random according to each other's positions in search of prey. Therefore, in this step the value of *A* takes values either greater than 1 or less than -1. In this step, the location of the search agents is determined randomly. The mathematical model of this situation is as in Equation 5 and Equation 6.

$$D = |C.X_{rand} - X| \tag{5}$$

$$X(t+1) = X_{rand} - A.D \tag{6}$$

High level features obtained from pre-trained networks are used in the framework of the proposed method. VGG-16 and ResNet-50 models were used as pre-trained network. These models were trained with the transfer learning method. For training and test data, the features obtained from pre-trained CNN networks were given as input to the designed ANN structure after the fusion process. Parameter update process in ANN model was performed with Whale Optimization Algorithm (WOA) instead of stochastic gradient descent algorithm. Weight and bias values have been updated more successfully than normal ANN models. The block diagram of the proposed approach is given in Figure 2.

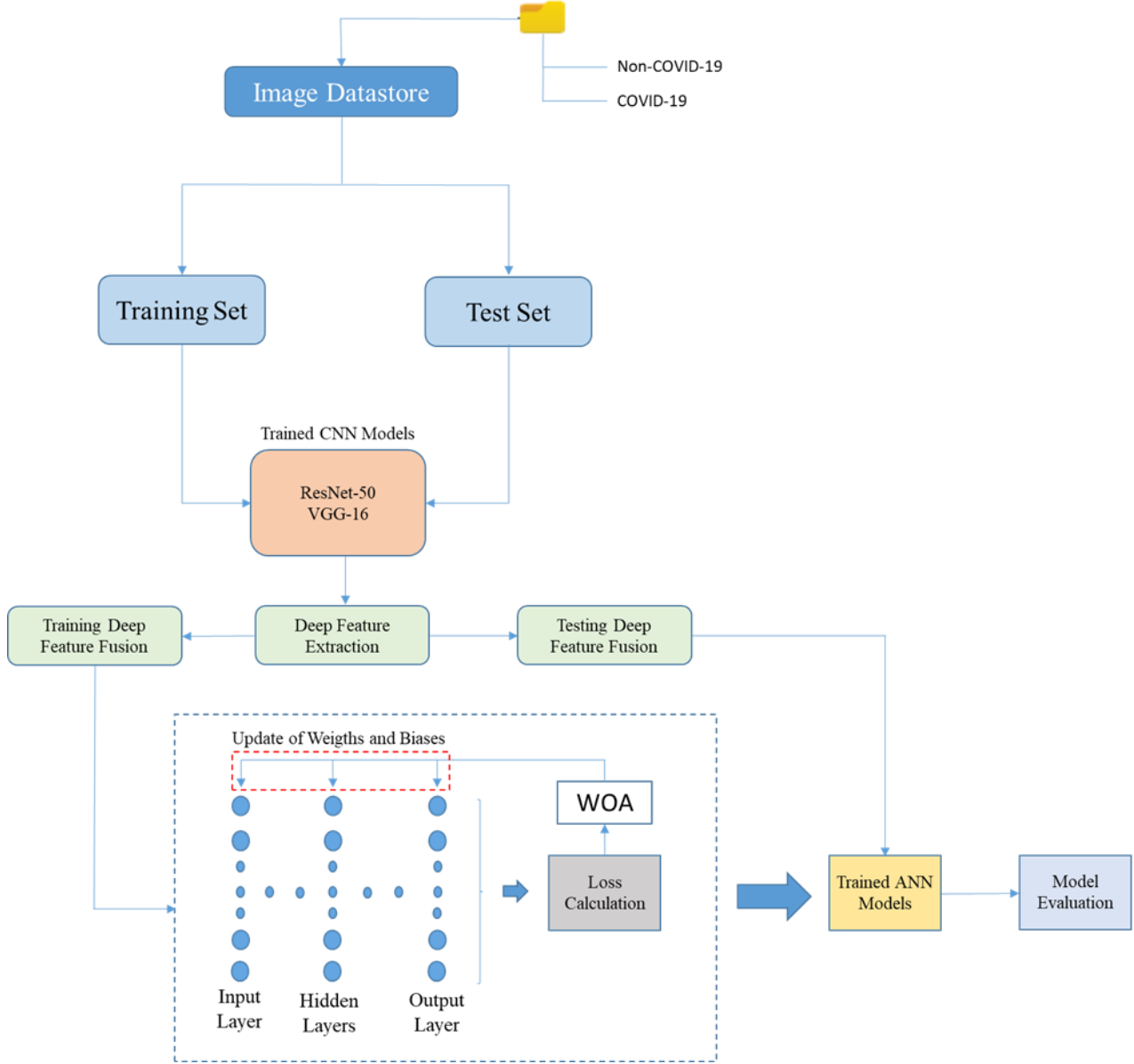

*Figure 2. The block diagram of the proposed model*

3. **RESULTS AND DISCUSSION**

As part of the study, the experiments were carried out on a workstation with Intel Core i7-7700 HQ CPU at 2.8 GHz, 16 GB RAM and NVIDIA GTX 1080 GPU. Matlab 2020a was preferred as a simulation environment.

Some performance metrics obtained from confusion matrix, which are Accuracy (ACC), Sensitivity (SEN), Specificity (SPE), F1-Score, Precision (PRE), Matthews Correlation Coefficient (MCC) and Kappa, were used to evaluate the models proposed in the study. These performance metrics are calculated by means of True Positive (TP), True Negative (TN), False Positive (FP) and False Negative (FN) indices as follows:

$$Accuracy = (TP+TN)/(TP+FN+TN+FP) \tag{7}$$

$$Sensitivity = TP/(TP+FN) \tag{8}$$

$$Specificity = TN/(TN+FP) \tag{9}$$

$$Precision = TP/(TP+FP) \tag{10}$$

$$F1-Score = (2 \times TP)/(2 \times TP + FN + FP) \tag{11}$$

$$MCC = \frac{TP \times TN - FP \times FN}{\sqrt{(TP+FP)(TP+FN)(TN+FP)(TN+FN)}} \tag{12}$$

$$Kappa = (total\ accuracy - random\ accuracy)/(1 - random\ accuracy) \tag{13}$$

By using transfer learning, training of GoogleNet, ResNet-50, DenseNet-201 and VGG-16 structures was carried out within pre-trained CNN models. During the training phase, the mini-batch size was determined as 10, and the training was completed in 30 epochs. The training process finished with 5580 iterations in total. Each epoch was completed after 186 iterations. The initial learning rate is 0.1. Learning drop factor and period are chosen as 0.1 and 10 epoch, respectively. Test data were used instead of validation data, and it was performed in each 1000 iterations. In this way, the optimization process can carried out with quickly and accurately. Stochastic Gradient Descent with Momentum (SGDM) method was preferred in the optimization process. Training and test charts of pre-trained CNN networks used in the study are available in Figure 3-6.

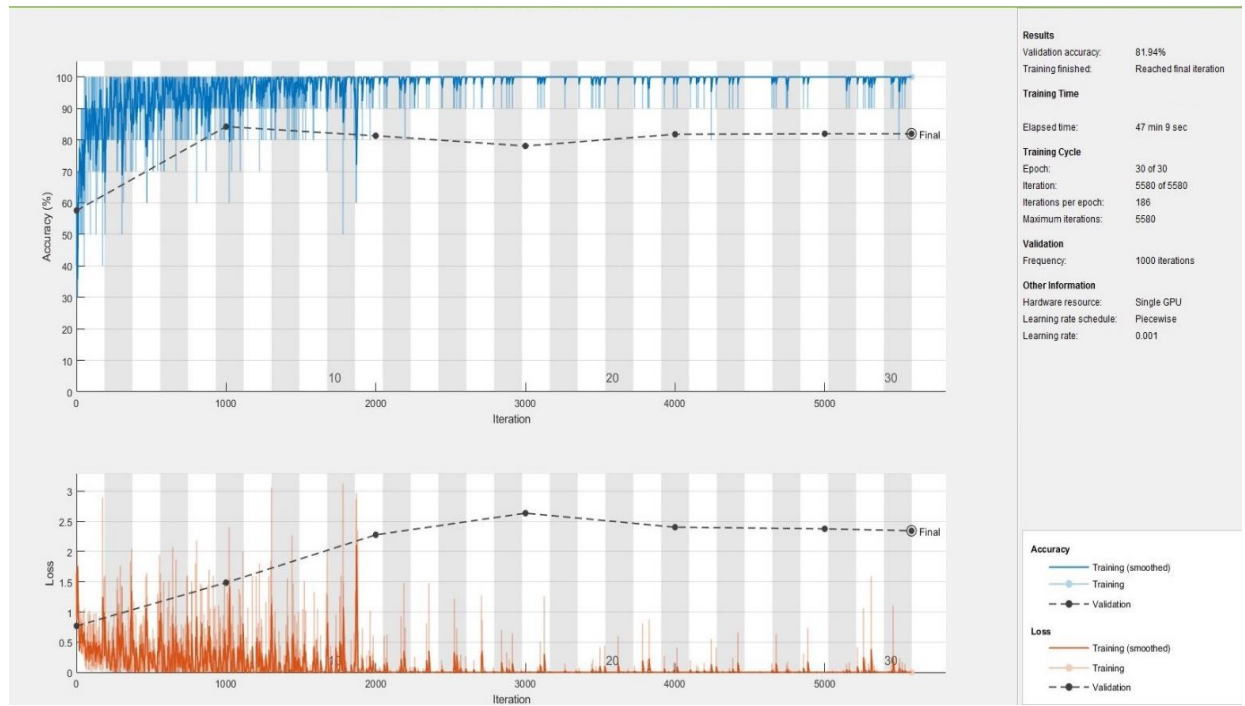

*Figure 3.* Training and Testing Process of GoogleNet

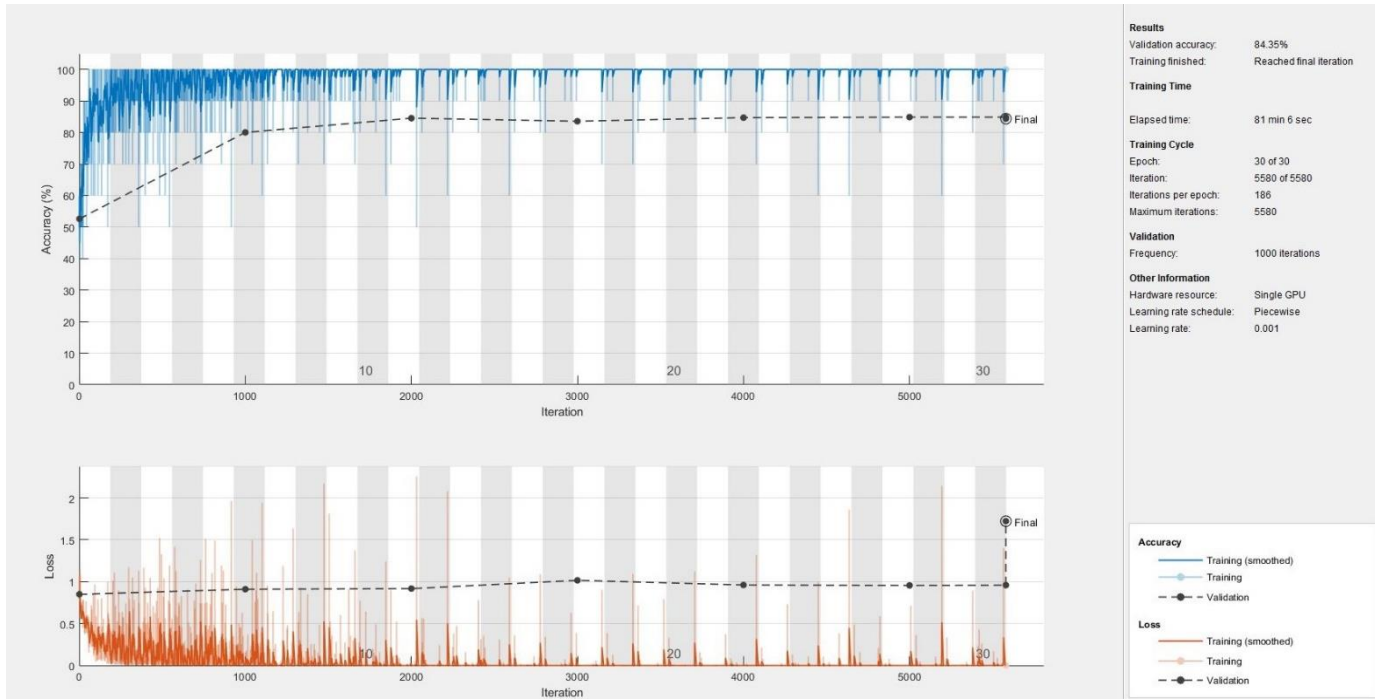

*Figure 4.* Training and Testing Process of ResNet-50

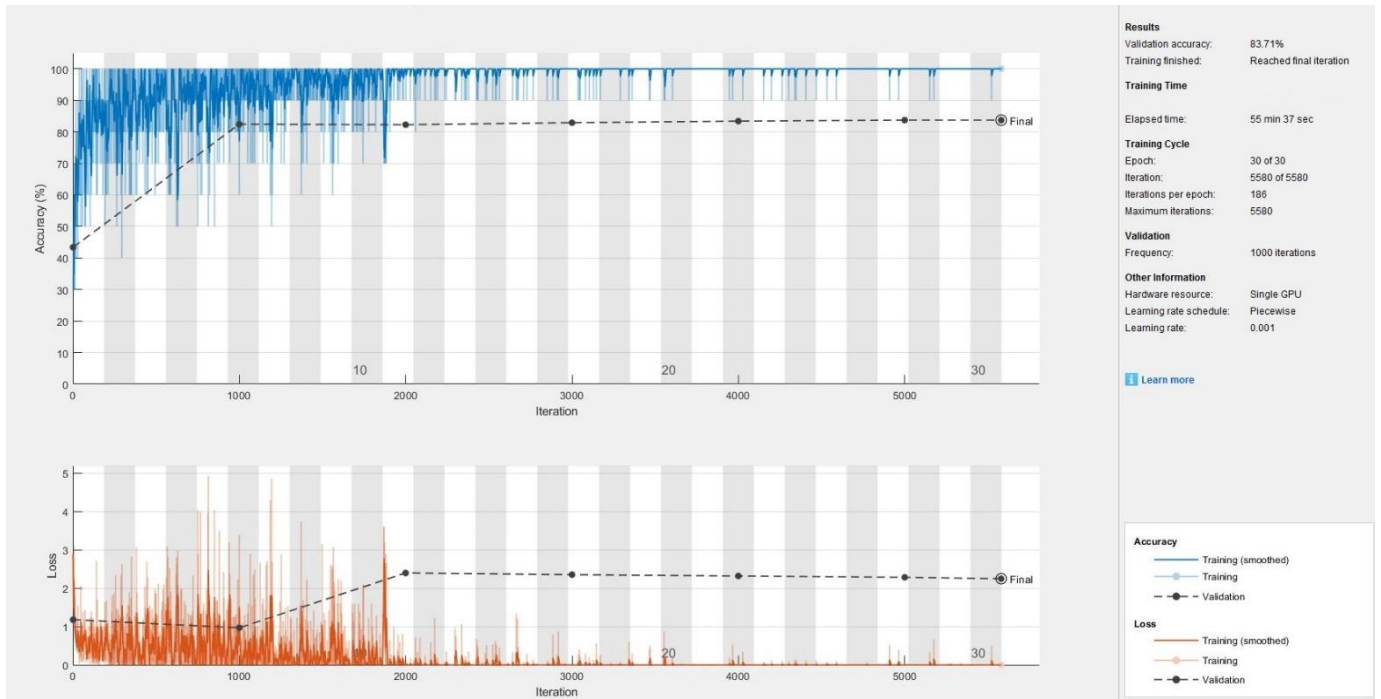

*Figure 5.* Training and Testing Process of VGG-16

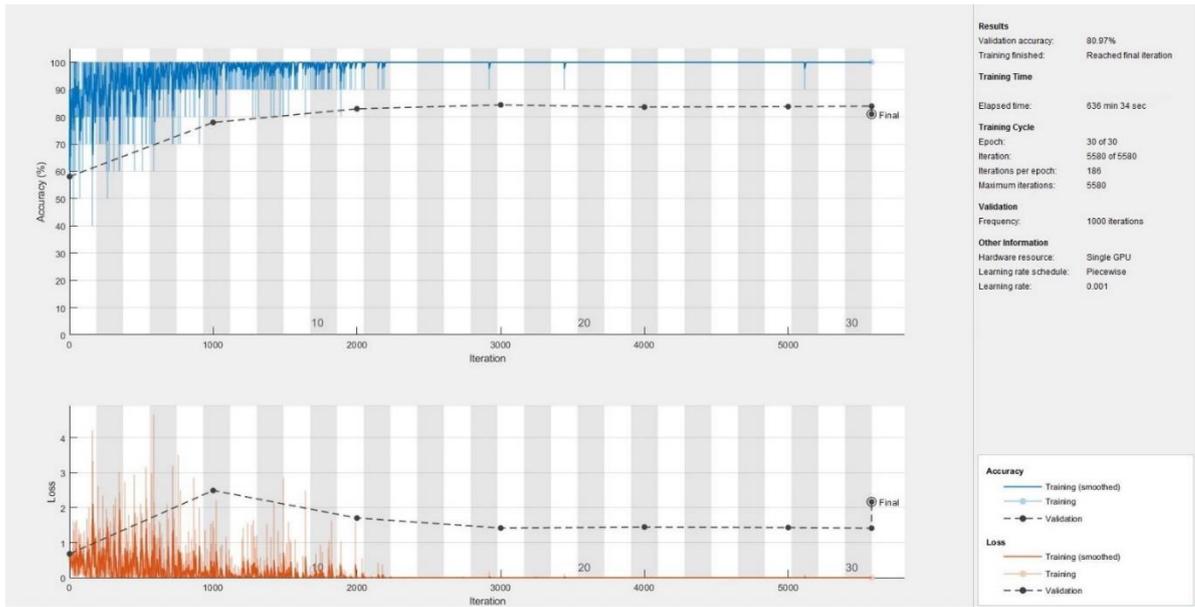

*Figure 6.* Training and Testing Process of DenseNet-201

There are two graphs in the figures, namely accuracy and loss. The blue curves on the accuracy graph represent the accuracy of the training data, while the black curves show the accuracy of the test data. The red curves in the loss graph represent the loss of training data, while the black curves represent the loss of test data. Since the number of parameters included in the pre-trained CNN models are different from each other, the training durations are also different from each other. The pre-trained CNN network with the shortest training period is GoogleNet and the training process was completed in 47 min 9 seconds. Then, VGG-16 model completed training with 55 min 37 seconds. There are two pre-trained CNN network models with the longest training period. One of them is ResNet-50 model and it completed the training process in 81 min 6 sec. DenseNet-201 is the model that has by far more training time than the training of other pre-trained CNN models. This model completed the training process in 636 min 34 sec. Achieving high classification accuracy in such short training times depends on transfer learning. In addition to the accuracy metric, other classification metrics are given in Table 1.

**Table 1.** Comparison of Metric Performance

| Methods | Evaluation Metrics (%) | | | | | | |
|---|---|---|---|---|---|---|---|
|  | ACC | SEN | SPE | PRE | F1-score | MCC | Kappa |
| VGG-16 | 83.71 | 68.05 | 99.67 | 99.53 | 80.83 | 71.22 | 67.52 |
| ResNet-50 | 84.35 | 69.01 | **100** | **100** | 81.66 | 72.42 | 68.80 |
| GoogleNet | 81.94 | 65.81 | 98.37 | 97.63 | 78.63 | 67.73 | 63.98 |
| DenseNet-201 | 80.97 | 62.62 | 99.67 | 99.49 | 76.86 | 66.89 | 62.07 |
| Proposed Method | **88.06** | **81.47** | 94.79 | 94.10 | **87.33** | **76.87** | **76.16** |

Table 1 includes the classification performance of four pre-trained CNN models and the proposed method. When evaluated in terms of performance, DenseNet-201 model showed the lowest performance. The performance of this model is 80.91% ACC, 62.62% SEN, 99.67% SPE, 99.49% PRE, 76.86% F1-Score, 66.89% MCC and 62.07% Kappa. ResNet-50 model showed the highest performance in pre-trained CNN models. The performance of the VGG-16 model is also very close to that of the ResNet-50 model. The classification metrics obtained with ResNet-50 were 84.35% ACC, 69.01% SEN, 100.00% SPE, 100.00% PRE, 81.66% F1-Score, 72.42% MCC and 68.80% Kappa. The proposed method has achieved higher performance than pre-trained CNN models. The performance of the proposed method is 88.06% ACC, 81.47% SEN, 94.79% SPE, 94.10% PRE, 87.33% F1-Score, 76.87% MCC and 76.16% Kappa. The proposed method stands out in all classification metrics except specitivity and precision metrics. In Figure 7, the confusion matrix of pre-trained CNN models is given.

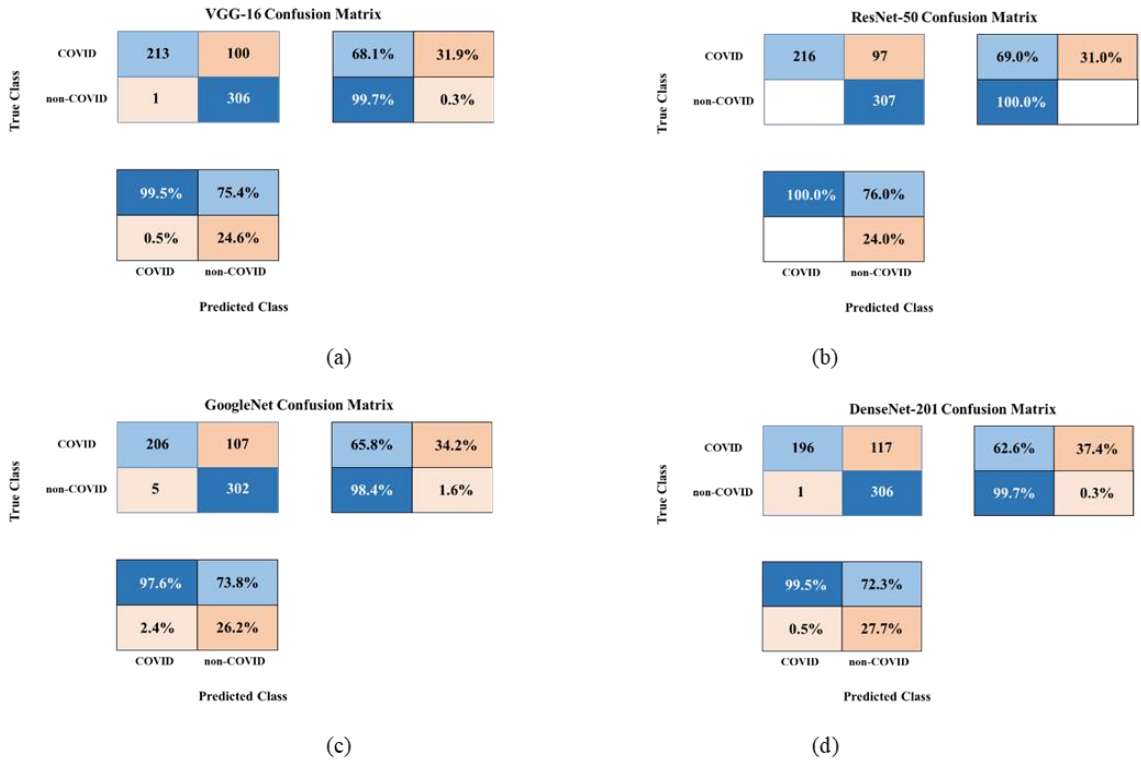

*Figure 7.* Confusion Matrix of pre-trained CNNs (a) VGG-16 (b) ResNet-50 (c) GoogleNet (d) DenseNet-201

Class accuracy and error rates can be seen in the confusion matrices obtained for test data. In the confusion matrix of the ResNet-50 model, 100% classification accuracy has been achieved in the non-COVID class. In the COVID class, this accuracy is 69.0%. In the DenseNet-201 model with the lowest performance, the classification accuracy achieved in the COVID class is 62.6%. In the non-COVID class, the classification accuracy is 99.7%. The highest error rate for the COVID class belongs to the DenseNet-201 model with 37.4%. Figure 8 contains the confusion matrix of the proposed method.

**Confusion Matrix of Proposed Method**

|  | COVID | non-COVID |  |  |
|---|---|---|---|---|
| COVID | 255 | 58 | 81.5% | 18.5% |
| non-COVID | 16 | 291 | 94.8% | 5.2% |

|  |  |
|---|---|
| 94.1% | 83.4% |
| 5.9% | 16.6% |
| COVID | non-COVID |

True Class / Predicted Class

*Figure 8.* Confusion Matrix of Proposed Method

In the concept of class accuracies of the proposed method, the accuracy rate has increased further in the COVID class. In the COVID class, the number of TN is 255 and the number of FP is 58. For the COVID class, the accuracy rate is 81.5%, and the error rate is 18.5%. For the Non-COVID class, the FN number is 16 and the number of TP is 291. Class accuracy rate is 94.8%, error rate is 5.2%.

4. **CONCLUSION**

Despite the COVID-19 outbreak, people continue their routine daily life. With the arrival of the winter months, the affected by this epidemic begin to rise significantly. Within the scope of this study, a COVID-19 diagnosis system was proposed to support clinical research. It is based on CNN architecture and utilized transfer learning. The deep learning method has automatic feature decoding capability and provides higher performance than handmade feature extraction engines. In this way, COVID-19 cases can be detected with a high performance by using CT images. The method used to achieve 88.06% ACC, 81.47% SEN, 94.79% SPE, 94.10% PRE, 87.33% F1-Score, 76.87% MCC and 76.16% Kappa classification performance. When the proposed method is evaluated clinically, it is predicted that it can support the decision-making process of experts. With this method, it has been proven that the error of misdiagnosis can be reduced, and the test times can be reduced from days.

**REFERENCES**

Afshar, P., Heidarian, S., Naderkhani, F., Oikonomou, A., Plataniotis, K. N., & Mohammadi, A. (2020). Covid-caps: A capsule network-based framework for identification of covid-19 cases from x-ray images. arXiv preprint arXiv:2004.02696.

Albahri O.S., Zaidan A.A., Albahri A.S.,. Zaidan B.B, Abdulkareem K. H., Al-qaysi Z.T., Alamoodi A.H., Aleesa A.M., Chyad M.A., Alesa R.M., Kem L.C., Lakulu M. M., Ibrahim A.B., Rashid N. A. (2020). Systematic review of artificial intelligence techniques in the detection and classification of COVID-19 medical images in terms of evaluation and benchmarking: Taxonomy analysis, challenges, future solutions and methodological aspects. Journal of Infection and Public Health, 13 (10), 1381-1396.

Barstugan, M., Ozkaya, U., & Ozturk, S. (2020). Coronavirus (covid-19) classification using ct images by machine learning methods. arXiv preprint arXiv:2003.09424.

Elaziz, M. A., Hosny, K. M., Salah, A., Darwish, M. M., Lu, S., & Sahlol, A. T. (2020). New machine learning method for image-based diagnosis of COVID-19. Plos one, 15(6), e0235187.

Fan, D. P., Zhou, T., Ji, G. P., Zhou, Y., Chen, G., Fu, H., ... & Shao, L. (2020). Inf-Net: Automatic COVID-19 Lung Infection Segmentation from CT Images. IEEE Transactions on Medical Imaging.

He, K., Zhang, X., Ren, S., & Sun, J. (2016). Deep residual learning for image recognition. In Proceedings of the IEEE conference on computer vision and pattern recognition (pp. 770-778).

Hemdan, E. E. D., Shouman, M. A., & Karar, M. E. (2020). Covidx-net: A framework of deep learning classifiers to diagnose covid-19 in x-ray images. arXiv preprint arXiv:2003.11055.

Jaiswal, A., Gianchandani, N., Singh, D., Kumar, V., & Kaur, M. (2020). Classification of the COVID-19 infected patients using DenseNet201 based deep transfer learning. Journal of Biomolecular Structure and Dynamics, 1-8.

Mirjalili, S., & Lewis, A. (2016). The whale optimization algorithm. Advances in engineering software, 95, 51-67.

Nour, M., Cömert, Z., & Polat, K. (2020). A novel medical diagnosis model for COVID-19 infection detection based on deep features and Bayesian optimization. Applied Soft Computing, 106580.

Pereira R. M., Bertolini D., Teixeira L. O., Silla C. N., Costa Y. M.G. (2020). COVID-19 identification in chest X-ray images on flat and hierarchical classification scenarios. Computer Methods and Programs in Biomedicine, 194.

Pham, T.D. (2020). A comprehensive study on classification of COVID-19 on computed tomography with pretrained convolutional neural networks. Nature, Sci Rep 10, 16942.

Randhawa, G. S., Soltysiak, M. P., El Roz, H., de Souza, C. P., Hill, K. A., & Kari, L. (2020). Machine learning using intrinsic genomic signatures for rapid classification of novel pathogens: COVID-19 case study. Plos one, 15(4), e0232391.

Sahlol, A. T., Yousri, D., Ewees, A. A., Al-Qaness, M. A., Damasevicius, R., & Abd Elaziz, M. (2020). COVID-19 image classification using deep features and fractional-order marine predators algorithm. Scientific Reports, 10(1), 1-15.

Shi, F., Xia, L., Shan, F., Wu, D., Wei, Y., Yuan, H., ... & Shen, D. (2020). Large-scale screening of covid-19 from community acquired pneumonia using infection size-aware classification. arXiv preprint arXiv:2003.09860.

Simonyan, K., & Zisserman, A. (2014). Very deep convolutional networks for large-scale image recognition. arXiv preprint arXiv:1409.1556.

Singh, D., Kumar, V., Yadav, V., & Kaur, M. (2020). Deep Neural Network-Based Screening Model for COVID-19-Infected Patients Using Chest X-Ray Images. International Journal of Pattern Recognition and Artificial Intelligence, 2151004.

Soares, E., Angelov, P., Biaso, S., Froes, M. H., & Abe, D. K. (2020). SARS-CoV-2 CT-scan dataset: A large dataset of real patients CT scans for SARS-CoV-2 identification. medRxiv.

Sun, L., Mo, Z., Yan, F., Xia, L., Shan, F., Ding, Z., ... & Yuan, H. (2020). Adaptive feature selection guided deep forest for covid-19 classification with chest ct. IEEE Journal of Biomedical and Health Informatics.

Ozturk, T., Talo, M., Yildirim, E. A., Baloglu, U. B., Yildirim, O., & Acharya, U. R. (2020). Automated detection of COVID-19 cases using deep neural networks with X-ray images. Computers in Biology and Medicine, 103792.

Öztürk, Ş., & Özkaya, U. (2020). Gastrointestinal tract classification using improved LSTM based CNN. Multimedia Tools and Applications, 1-16.

Öztürk, Ş., Özkaya, U., & Barstuğan, M. (2020). Classification of Coronavirus (COVID-19) from X-ray and CT images using shrunken features. International Journal of Imaging Systems and Technology.

Ucar, F., & Korkmaz, D. (2020). COVIDiagnosis-Net: Deep Bayes-SqueezeNet based Diagnostic of the Coronavirus Disease 2019 (COVID-19) from X-Ray Images. Medical Hypotheses, 109761.

Vaishya, R., Javaid, M., Khan, I. H., & Haleem, A. (2020). Artificial Intelligence (AI) applications for COVID-19 pandemic. Diabetes & Metabolic Syndrome: Clinical Research & Reviews.

Wang X., Deng X., Fu Q., Zhou Q., Feng J., Ma H., Liu W., and Zheng C. (2020). A Weakly-Supervised Framework for COVID-19 Classification and Lesion Localization From Chest CT. IEEE Transactions on Medical Imaging, 39(8) , 2615-2625.

Wu, Z., Ling, Q., Chen, T., & Giannakis, G. B. (2020). Federated variance-reduced stochastic gradient descent with robustness to byzantine attacks. IEEE Transactions on Signal Processing, 68, 4583-4596.

Zu, Z. Y., Jiang, M. D., Xu, P. P., Chen, W., Ni, Q. Q., Lu, G. M., & Zhang, L. J. (2020). Coronavirus disease 2019 (COVID-19): a perspective from China. Radiology, 200490.